%% file: arxiv.tex
\documentclass[conference]{ieeeconf}
\IEEEoverridecommandlockouts
\input{includes}

\usepackage{amssymb}

\title{\LARGE \bf
\method: Meta Learning and Planning for Nonprehensile \\ Manipulation of Unseen Objects with Rapid Online Adaption
}

\author{Donghyung~Lee$^*$,
        Seyedali~Golestaneh$^*$,
        Jaskrit~Singh,
        Zhuoyun~Zhong,
        \\Athanasios~Kapoutsis,
        and Constantinos~Chamzas
}

\begin{document}

\maketitle
\thispagestyle{empty}
\pagestyle{empty}


\renewcommand\twocolumn[1][]{#1}%
\maketitle

\input{0.Abstract}
\input{1.Introduction}
\input{2.RelatedWork}
\input{3.ProblemStatement}
\input{4.Methodology}
\input{5.Experiments}
\input{6.Conclusion}


\bibliographystyle{IEEEtran_ShortURL}
\bibliography{reference}

\end{document}

%% file: includes.tex
\usepackage[utf8]{inputenc}
\usepackage{comment}
\usepackage{graphicx}
\usepackage{xcolor}
\usepackage{wrapfig}
\usepackage{lipsum}
\usepackage[dvipsnames]{xcolor}
\definecolor{rred}{RGB}{169, 50, 38}
\definecolor{ggreen}{RGB}{34, 153, 84}
\definecolor{bblue}{RGB}{36, 113, 163}
\definecolor{ppurple}{RGB}{125, 60, 152}
\definecolor{yyellow}{RGB}{214, 137, 16}
\definecolor{ggrey}{RGB}{112, 123, 124}

\usepackage{amsmath}
\usepackage{amssymb}
\usepackage{amsthm}
\let\labelindent\relax
\usepackage{enumitem}
\theoremstyle{definition}
\usepackage{tikz}
\usetikzlibrary{automata,positioning,angles,quotes}
\usepackage{tikz-cd}

\newtheoremstyle{main}
{1em} 
{1em} 
{\normalfont} 
{0pt} 
{\scshape} 
{\\*} 
{2pt} 
{\thmname{#1}\thmnumber{ #2}: \thmnote{\itshape #3}} 

\makeatletter
\let\NAT@parse\undefined
\makeatother
\usepackage{cite}
\usepackage[pdfa,colorlinks,bookmarks=true,bookmarksopen,bookmarksnumbered,allcolors=ggreen]{hyperref}

\usepackage[algoruled,vlined,linesnumbered]{algorithm2e}

\SetAlgoSkip{SkipBeforeAndAfter}

\SetCommentSty{mycommfont}
\SetKw{Continue}{continue}
\SetAlgorithmName{Algorithm}{Alg.}

\usepackage{booktabs}
\usepackage{multirow}
\usepackage{multicol}
\usepackage{makecell}

\usepackage[font=footnotesize]{caption}
\usepackage[font=footnotesize]{subcaption}
\usepackage[export]{adjustbox}

\usepackage[
activate   = {true},
protrusion = true,
expansion  = true,
kerning    = true,
spacing    = true,
tracking   = false,
auto       = true,
selected   = true,
factor     = 1000,
stretch    = 20,
shrink     = 20,
]{microtype}

\usepackage{xspace}

\newcommand{\method}{\textsc{MetaPusher}\xspace}
\newcommand{\aorrt}{\textsc{AORRT}\xspace}
\newcommand{\osm}{\textsc{Object-specific Model}\xspace}
\newcommand{\pft}{\textsc{Pooled Fine-tuning}\xspace}
\newcommand{\ml}{\textsc{Meta-learning}\xspace}
\newcommand{\al}{\textsc{Active Learning}\xspace}
\newcommand{\hm}{\textsc{HACMan}\xspace}
\newcommand{\ap}{\textsc{Adaptive Planning}\xspace}

\newcommand{\start}{\text{start}}
\newcommand{\goal}{\text{goal}}
\newcommand{\free}{\text{free}}
\newcommand{\train}{\text{train}}

\newcommand{\test}{\text{test}}
\newcommand{\obs}{\text{obs}}
\newcommand{\gt}{\text{gt}}

\newcommand{\SEThree}{\ensuremath{SE(3)}\xspace}
\newcommand{\SETwo}{\ensuremath{SE(2)}\xspace}

\newcommand{\X}{\ensuremath{\mathcal{X}}\xspace}
\newcommand{\U}{\ensuremath{\mathcal{U}}\xspace}
\newcommand{\T}{\ensuremath{\mathcal{T}}\xspace}

\newcommand{\D}{\ensuremath{\mathcal{D}}\xspace}

\newcommand{\Objs}{\ensuremath{\mathcal{O}}\xspace}


%% file: 0.Abstract.tex
\begin{abstract}
    Manipulating previously unseen objects remains challenging, as their dynamics depend on latent physical properties, such as friction and mass distribution, that cannot be inferred from perception alone.
    Prior experience across objects can provide an initial estimate of unseen object dynamics, but this estimate remains uncertain and can degrade further during sim-to-real transfer.
    Adapting the dynamics through interaction can progressively refine the estimation, however, updating the model may invalidate the planned trajectory.
    Successful and efficient manipulation therefore requires both rapid dynamics adaptation and a planning strategy that can incorporate this evolution.
    In this work, we introduce \method, a meta-learning and adaptive planning framework for nonprehensile manipulation of unseen objects without prior object-specific interactions.
    A meta-learned dynamics model rapidly adapts from interactions during task execution, while an adaptive kinodynamic planner updates long-horizon plans by reusing and refining its existing search tree.
    %
    This coupling enables manipulation and adaptation without a separate data collection phase.
    %
    We evaluate \method on unseen objects in simulation and in sim-to-real scenarios, comparing against fine-tuning and active learning methods, MPPI-based control, and a reinforcement learning policy. It achieves lower prediction error and improves task success rate by up to \textbf{20\%}. The source code is available at \url{https://anonymous.4open.science/r/metapusher-B2BB/README.md}
    %
\end{abstract}

%% file: 1.Introduction.tex
\section{Introduction}
\label{sec:introduction}
%
%
Nonprehensile manipulation of previously unseen objects is challenging because pushing outcomes depend not only on the object's geometry but also on its latent physical properties, such as friction, mass distribution, and inertia. Analytical models provide useful structure for planar pushing~\cite{stable-pushing, mechanics-of-push}, while data-driven methods learn the dynamics from collected interaction data~\cite{activepusher, data-efficient-pushing, learn-poke-by-poking}. However, in a more general setting, when a robot encounters a new object, it should be able to begin manipulation without prior object-specific interactions.

Existing methods address parts of this problem separately. Generalizable manipulation methods~\cite{hacman, corn} transfer experience across objects from geometric features, but this information alone cannot capture the full object behavior. In contrast, adaptive methods update dynamics from interactions~\cite{grbal, hyperdynamics, push-net, dywa} but typically rely on dedicated data collection and short-horizon control. Kinodynamic planning methods~\cite{sbmp, kraft, aura} provide long-horizon reasoning, but their trajectories are generated under a fixed forward model and may no longer be dynamically feasible when that model evolves. Therefore, efficiently manipulating unseen objects requires adapting both the dynamics model and the plan during task execution.

In this paper, we introduce \method, a meta-learning and adaptive planning framework for manipulation of unseen objects with no prior interactions. \method uses MAML\cite{maml} to learn an adaptive dynamics model across diverse objects. At deployment, an adaptive kinodynamic planner uses the dynamics estimate to generate a long-horizon trajectory. After each executed push, the dynamics model adapts based on the recently observed transitions, and the adaptive kinodynamic planner prunes and refines the existing search tree under the updated model rather than replanning from scratch. This couples online dynamics adaptation with long-horizon adaptive planning and requires no separate object-specific data collection phase.

\begin{figure}
    \centering
    \includegraphics[width=1\linewidth]{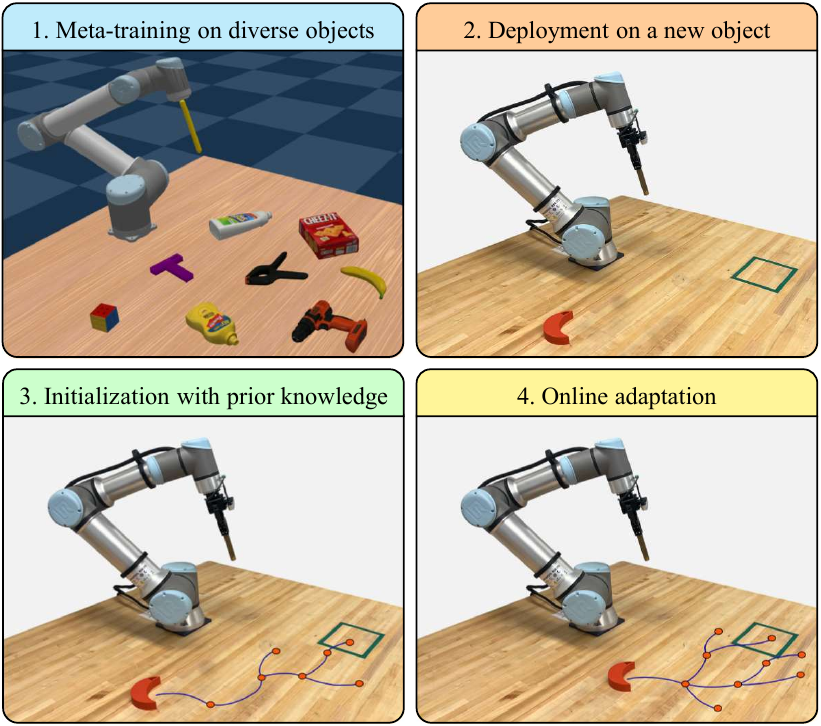}
    \caption{\textbf{\method Overview: \textcolor{TealBlue}{(1)}} An initial forward dynamics model is meta-trained across diverse objects in simulation; \textbf{\textcolor{YellowOrange}{(2)}} At deployment, a previously unseen object is encountered for the first time, with zero prior interactions; \textbf{\textcolor{LimeGreen}{(3)}} The manipulation task begins directly using the initial dynamics model; \textbf{\textcolor{Goldenrod}{(4)}} Executed pushes adapt both the dynamics model and the existing planning tree under the evolving dynamics.}
    \vspace{-1em}
\end{figure}

Our main contributions are:
\begin{itemize}
    \item A geometry-conditioned meta-learning dynamics model that enables manipulation of unseen objects without prior object-specific interactions and rapidly adapts from recent transitions during task execution.
    \item An adaptive kinodynamic planning strategy that retains and updates the existing search tree as the dynamics model updates.
    \item Extensive simulation and real-world evaluation across a total of 14 unseen objects, demonstrating effective generalization and sim-to-real transfer, including a \textbf{100\%} task success rate in the real world.
\end{itemize}

%% file: 2.RelatedWork.tex
\section{Related Work}
\label{sec:relatedWork}
This section reviews the prior work most relevant to our method from three perspectives: (i)~generalizable nonprehensile manipulation, (ii)~online dynamics adaptation, (iii)~long-horizon kinodynamic planning.

\subsection{Generalizable Nonprehensile Manipulation}
Analytical models of nonprehensile pushing have been developed to enable planar manipulation planning~\cite{stable-pushing, mechanics-of-push}. However, in practice, accurate prediction remains difficult since pushing outcomes depend on geometry, contact, friction, and mass distribution~\cite{data-efficient-pushing}. Data-driven methods therefore learn planar pushing dynamics directly from interaction data~\cite{million-ways-to-push,learn-poke-by-poking}, while methods like ActivePusher~\cite{activepusher} improve data efficiency through uncertainty-guided data acquisition. Such approaches can provide accurate models, but require dedicated object-specific interaction data to characterize the dynamics of an unseen object.

A complementary line of work learns across diverse objects using simulation interactions and transfers prior experience to previously unseen objects using visual or geometric representations. HACMan~\cite{hacman, hacman-plusplus}, CORN~\cite{corn} learn point-cloud-based policies for zero-shot sim-to-real transfer and unseen object generalizability, while Push~Anything~\cite{push-anything} uses contact implicit MPC using geometry reconstruction. Although these methods generalize well to unseen objects, observable geometry alone cannot capture latent physical properties that govern object motion. In contrast, \method begins the manipulation directly from geometric observations and progressively adapts the dynamics online using interactions generated during task execution, without requiring a separate object-specific data collection phase.

\subsection{Online Dynamics Adaptation}
Online adaptation estimates dynamics model information during interaction. Meta-learning enables rapid specialization from limited experience, with MAML~\cite{maml} learning an initialization that can be adapted with only a few gradient updates. GrBAL~\cite{grbal} exploits this principle and learns a meta-learned dynamics that updates from recent transitions, while \cite{off-road-meta} meta-learns parameters for real-time online dynamics adaptation in off-road autonomous driving. HyperDynamics~\cite{hyperdynamics} specializes object dynamics after acquiring a small number of interactions with a new object before starting the manipulation. Push-Net~\cite{push-net} recurrently uses pushing history to reason about unknown physical properties, RMPPI~\cite{rmppi} uses recurrent forward dynamics to adapt simulation-trained models to real object behavior, and DyWA~\cite{dywa} combines manipulation history with point cloud geometry for sim-to-real adaptation. Unlike prior methods that primarily require separate interactions, \method learns an adaptive forward dynamics model based on geometry, begins goal directed manipulation with zero object specific data, and specializes the model through task interactions. Moreover, instead of short-horizon control or repeated MPC~\cite{mpc}, \method incorporates the evolving dynamics into long-horizon kinodynamic planning. This enables instant manipulation of unseen objects while adapting both the dynamics model and the plan.

\subsection{Long-Horizon Kinodynamic Planning}
Most adaptive dynamics approaches above employ receding-horizon optimization, including MPC and MPPI variants~\cite{mppi}. Although these controllers can incorporate updated dynamics at each cycle, their finite-horizon optimization can lead to myopic behavior. Sampling-based kinodynamic planners~\cite{sbmp} instead search directly over dynamically feasible trajectories and have been applied to long-horizon manipulation, as in \cite{activepusher}. KRAFT~\cite{kraft} and AURA~\cite{aura} reuse prior planning information during execution to improve robustness and replanning efficiency. Meanwhile, recent GPU-parallelized planners, such as Kino-PAX~\cite{kinopax} and PAKR~\cite{pakr}, reduce planning time substantially through parallel tree expansion. These methods, however, assume a fixed dynamics model during planning and do not address the case where the forward model changes. \method instead couples online dynamics adaptation with long-horizon kinodynamic planning by retaining and updating the existing search tree as the learned dynamics evolve, rather than planning from scratch under a fixed model.
\begin{figure*}[t]
    \centering
    \includegraphics[width=\textwidth]{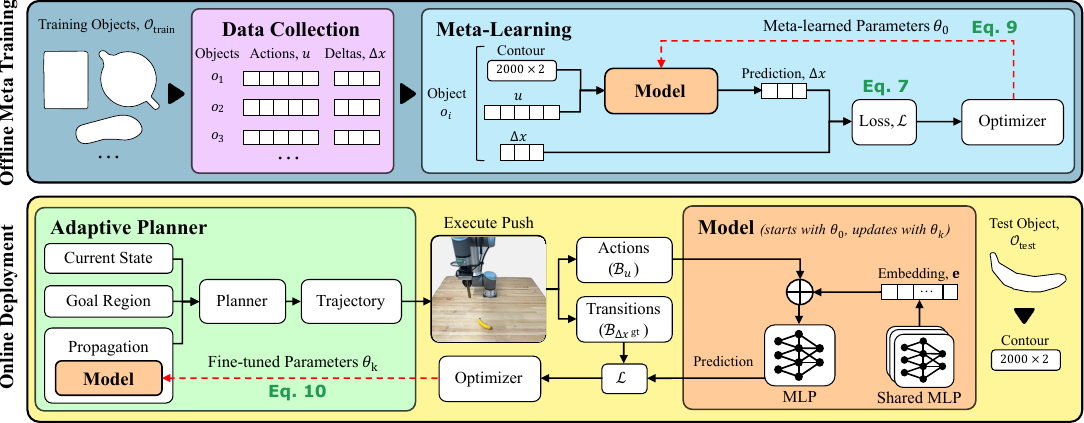}
    \caption{\textbf{\method Framework.} During \textit{\textcolor{NavyBlue}{Offline Meta Training}}, interaction data from training objects are used to learn the initialization $\theta_0$ of the geometry-informed forward dynamics model. At \textit{\textcolor{Dandelion}{Online Deployment}}, the planner uses the current dynamics model to generate a trajectory and execute a push. The resulting transition is then used to update the model parameters to $\theta_k$. The adapted model is then returned to the planner to update the search tree, allowing the dynamics model and planned trajectory to evolve together while manipulating a previously unseen object.}
    \label{fig:overview}
    \vspace{-1em}
\end{figure*}

%% file: 3.ProblemStatement.tex
\section{Problem Statement}
\label{sec:problem}
We consider planar pushing manipulation of a previously unseen object with unknown dynamics. Let $\X$ and $\U$ denote the object's state space and control space, respectively. The state space $\X$ is partitioned into two disjoint subsets, \mbox{$\X = \X_\obs \;\cup\; \X_\free$}, where $\X_{\text{obs}}$ represents the invalid state region and $\X_{\text{free}}$ is the valid state region.

Let $\Objs_\train$ denote the set of objects available during training and $\Objs_\test$ the set used for evaluation, where \mbox{$\Objs_\train \cap \Objs_\test = \varnothing$}. For each $o \in \Objs_\train$, arbitrary interaction data $\D^\train_o = \{(x_j,u_j,x_{j+1})\}_{j=1}^M$ may be collected during training in simulation, whereas each \mbox{$o\in\Objs_\test$} is encountered for the first time at deployment with zero prior interaction data $\D^\test_o = \varnothing$. The result of a push on object $o$ is governed by an unknown transition function:
\begin{equation}
\label{eq:ground-truth-dynamics}
   x_{k+1}^\gt =\Gamma_o^\gt (x_k^\gt, u_k),
\end{equation}
where $x_{k}^\gt$ is the state at iteration $k$, and $x_{k+1}^\gt$ is the ground-truth state after applying $u_k$.

\textit{Problem Definition:} For an object $o \in \Objs_\test$ with an initial state $x_\start \in \X_\free$, the manipulation objective is to find the shortest feasible trajectory \mbox{$\pi=(x_0, u_0, \ldots, x_{T-1}, u_{T-1}, x_T)$} that drives the object to the goal region $\X_\goal \subseteq \X_\free$, such that:
\begin{equation}
\label{eq:problem}
\begin{gathered}
x_{i+1} = \Gamma_o^\gt(x_i,u_i), \qquad x_0 = x_\start, \qquad x_T \in \X_\goal, \\
x_{0:T} \in \X_\free, \qquad u_{0:T-1} \in \U.
\end{gathered}
\end{equation}

In the next section, we present \method, our approach for addressing this problem.

%% file: 4.Methodology.tex
\section{Methodology}
\label{sec:methodology}
Our method consists of two phases: \textit{Offline Meta Training} and \textit{Online Deployment}. During offline training, interaction data $\D^\train_o$ from objects $o \in \Objs_\train$ are used to learn an initial geometry-encoded forward dynamics model (\autoref{subsec:meta-learning}). During online deployment, an object $o \in \Objs_\test$ is encountered with no prior interactions, and the initial forward dynamics model is used to start the manipulation. The model parameters are continuously adapted using each observed transition (\autoref{subsec:online-adaptation}), while an adaptive planning strategy updates the existing planning tree to remain consistent with the evolving dynamics (\autoref{subsec:adaptive-planning}). \autoref{fig:overview} summarizes the complete framework.

Push actions are defined from the object's local 2D contour point cloud, $g_o$. We select a contact position $[c_x,c_y]$ on the contour and apply the push along the inward surface normal at the contact point, with angle $\phi$. To avoid the angular discontinuity at $\pm\pi$, the pushing angle is represented by $[\sin\phi,\cos\phi]$. Although $c_x$ and $c_y$ fully determine $\phi$, we found explicitly including it to be beneficial for training. With a push distance \mbox{$d \in [d_{min}, d_{max}]$}, each push action is represented as:
\begin{equation}
\label{eq:push-params}
    u = [c_x, c_y, \sin\phi, \cos\phi, d].
\end{equation}

To incorporate object geometry into the dynamics model, $g_o$ is encoded using a PointNet-style architecture~\cite{point-net}, extracting an 1024-dimensional global feature vector $\mathbf{e} = [e_1, \dots, e_{1024}]$. This embedding provides a global representation of the object's contour geometry and is concatenated with the push action and provided directly to the forward dynamics model:
\begin{equation}
\label{eq:input}
    [c_x,\ c_y,\ \sin\phi,\ \cos\phi,\ d,\ e_1,\ \dots,\ e_{1024}]^\top \in \mathbb{R}^{1029}.
\end{equation}

Then, the learned forward dynamics model uses this concatenated representation to predict the resulting
planar object displacement, represented in $\SETwo$ as:
\begin{equation}
\label{eq:post-push}
    \Delta{x} = [\Delta p_x,\Delta p_y,\Delta\psi].
\end{equation}

\subsection{Offline Meta Training}
\label{subsec:meta-learning}
Each $o \in \Objs_\train$ defines an individual meta-learning objective. Using the corresponding interaction data from $\D^\train_o$, we seek a general initialization $\theta_0$ for the forward dynamics model $\Gamma^\theta$ that captures transferable pushing behavior to $\Objs_\test$.
Conditioned on the object geometry $g_o$ and push action, $\Gamma_\theta$ predicts the state transition:
\begin{equation}
\label{eq:prediction}
    x_{k+1}=\Gamma_{\theta}(x_k,u_k;g_o).
\end{equation}

The learning objective is a weighted mean squared error between the predicted and observed planar displacement:
\begin{equation}
\label{eq:loss-error}
    \mathcal{L} = \left\| \Delta p - \Delta p^\gt \right\|_2^2
    +
    w_\psi \left\| \Delta \psi - \Delta \psi^\gt
    \right\|_2^2,
\end{equation}
where $\Delta p=[\Delta p_x,\Delta p_y]^\top$ is the translational displacement, and $w_\psi = 0.2$ is the  weight of the rotational displacement error.

We train our model using a meta-learning framework, following Model-Agnostic Meta-Learning (MAML)~\cite{maml}. Offline Meta Training is performed across $\Objs_\train$ with only a small amount of data per object for adaptation, matching the deployment setting in which only limited interaction data become available for unseen objects.

For each training object $o_i$, a support set $\D_{o_i}^{\mathrm{support}}$ and a query set $\D_{o_i}^{\mathrm{query}}$ are sampled from the interaction data $\D_{o_i}^{\mathrm{train}}$ and used for per-object adaptation and evaluation of the adapted model within meta-learning, respectively, such that:
\begin{equation}
\label{eq:support_and_query}
\begin{gathered}
\D_{o_i}^{\mathrm{support}} \cup \D_{o_i}^{\mathrm{query}}
\subseteq \D_{o_i}^{\mathrm{train}}, \\
\D_{o_i}^{\mathrm{support}} \cap \D_{o_i}^{\mathrm{query}}
= \varnothing.
\end{gathered}
\end{equation}

The meta-learned parameters are then updated across the training objects as:
\begin{equation}
\label{eq:maml}
\theta \leftarrow \theta - \beta \sum_i
\nabla_\theta
\mathcal{L}
(
\underbrace{
\theta - \alpha \nabla_\theta
\mathcal{L}(\theta,\D_{o_i}^{\mathrm{support}})
}_{\substack{\theta'_{o_i}, \ \text{Inner Gradient Step}}},
\D_{o_i}^{\mathrm{query}}
),
\end{equation}
where $\alpha$ is the inner-loop adaptation learning rate, $\beta$ is the outer-loop meta adaptation learning rate, and $\mathcal{L(\theta,\D)}$ denotes the loss defined in \autoref{eq:loss-error}. The inner gradient step adapts the model parameters to each object using $\mathcal{D}_{o_i}^{\text{support}}$, while the outer loop evaluates the object-specific parameters on $\mathcal{D}_{o_i}^{\text{query}}$ and backpropagates the resulting loss to update the shared parameters $\theta$ accordingly. This explicitly encourages $\theta$ to serve as an initialization from which the model can rapidly adapt to the pushing dynamics of a new object.

\subsection{Online Adaptation}
\label{subsec:online-adaptation}
For a previously unseen object $o \in \Objs_\test$, deployment begins using $\Gamma_{\theta_0}(\cdot, \cdot)$ with no prior interactions. The initial dynamics model therefore relies on the parameters transferred from $\Objs_\train$ together with the observed object geometry $g_o$. After executing each push $u_k$, the action and the observed transition $\Delta x^\gt$ from $x_k^\gt$ to $x_{k+1}^\gt$ are stored on $\mathcal{B}_u$ and $\mathcal{B}_{\Delta x^\gt}$, respectively, and used to update the model parameters from $\theta_k$ to ${\theta_{k+1}}$.

Let $\D_{o,k}^\test$ denote the interaction data available for object $o$ after $k$ iteration, assuming $\D_{o,0}^\test=\varnothing$. From the current parameters $\theta_k$, online adaptation follows the same gradient update used in meta-learning of \autoref{eq:maml}, but is now performed using transitions generated during the ongoing manipulation task:
\begin{equation}
\label{eq:online-adaptation}
\theta_{k+1} = \underbrace{ \theta_k - \alpha \nabla_{\theta_k} \mathcal{L}
(\theta_k,\D_{o,k}^{\mathrm{test}})
}_{\text{Online Gradient Update}}.
\end{equation}
The updated model $\Gamma_{\theta_{k+1}}$ is then used by the adaptive planner to update the current trajectory, allowing the dynamics model and long-horizon trajectory to evolve together as additional interactions become available.

\subsection{Adaptive Planning}
\label{subsec:adaptive-planning}
\method uses a sampling-based kinodynamic planner to generate long-horizon dynamically feasible pushing trajectories under the adaptive dynamics model. At execution iteration $k$, given the current state $x_k^\gt$ and current dynamics $f^{\theta_{k}}_o$, the planner computes a trajectory \mbox{$\pi_k=(x_0, u_0, \ldots, x_{T-1}, u_{T-1}, x_T)$}, such as:
\begin{equation}
\label{eq:planning}
\begin{gathered}
    x_0 = x_k^\gt, \qquad x_T \in \X_\goal, \qquad x_{i+1} = x_i \oplus f_o^{\theta_k}(x_i, u_i), \\
    x_{0:T} \in \X_\free, \qquad u_{0:T-1} \in \U,
\end{gathered}
\end{equation}
where $\oplus$ denotes the pose composition operator in \SETwo.

After executing each control and observing the resulting state $x_{k+1}^\gt$, the online adaptation process, described in \autoref{subsec:meta-learning}, changes the dynamics model from $f^{\theta_k}_o$ to $f^{\theta_{k+1}}_o$. Consequently, the search tree propagated under the previous model may become inconsistent with the updated dynamics. Moreover, prediction discrepancies can accumulate through its descendants. To address this, the proposed adaptive planner retains and updates the existing search tree after each model update rather than replanning from scratch.
\begin{figure}[h]
    \centering
    \includegraphics[width=1\linewidth]{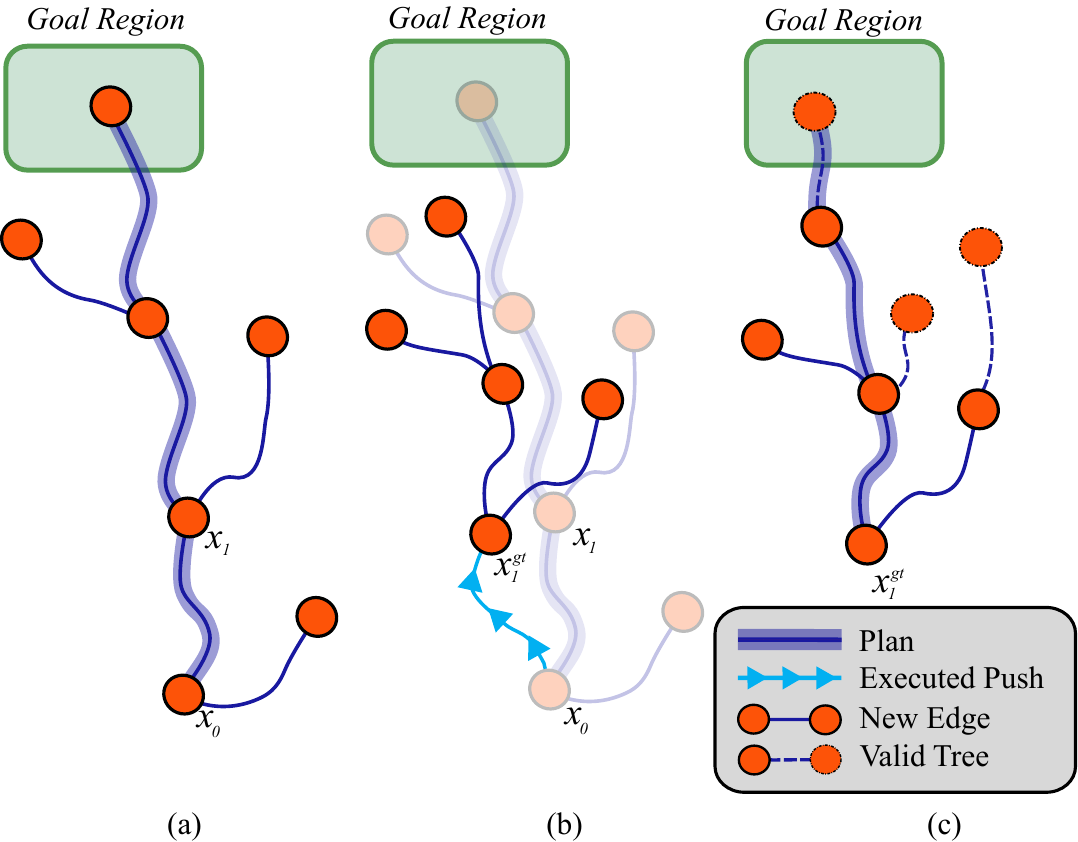}
    \caption{\textbf{Adaptive Planning.} \textbf{(a)} Initial planning tree searching for a path to the goal region. \textbf{(b)} After the first push, we arrive at $x_1^{gt}$ instead of $x_1$, so we update our pushing model accordingly. We use the new pushing model to reconstruct the old tree. \textbf{(c)} If the reconstructed tree no longer reaches the goal region, we expand the tree until a new path to the goal is found.}
    \label{fig:adaptive_planning}
    \vspace{-1em}
\end{figure}

\autoref{fig:adaptive_planning} shows the search tree available at execution iteration $k$, $\T_k$, where vertices denote predicted states and edges represent the push actions. After selecting and executing the next action $u_0$ from the current solution, the resulting state $x_{k+1}^{\gt}$ is observed and used to adapt the model to $f^{\theta_{k+1}}_o$. The planning tree is then adapted through three operations: \textit{Prune}, \textit{Refinement}, and \textit{Expansion}.

\textbf{Prune:} After executing $u_0$, the branches that are no longer descendants of $x_1$ are not dynamically reachable and are removed. This strategy retains a subtree that preserves search exploration that remains valid from the new state.

\textbf{Refinement:} The planner then sets the new root at the observed state $x_1^\gt$, and reconstructs the remaining subtree under $f^{\theta_{k+1}}_o$, proceeding from the root toward the leaves so that the updated plan is consistent through all the states. The planner checks the validity of vertices of the new tree $\T_{k+1}$ and removes any vertex propagated into $\X_\obs$.

\textbf{Expansion:} If the refinement invalidates the previous goal-reaching trajectory, the planner resumes the search from $\T_{k+1}$ and adds new nodes propagated using $f^{\theta_{k+1}}_o$. This allows \method to leverage previous exploration while improving solution accuracy under the updated dynamics estimate.

This process is repeated after each object interaction. Therefore, \method couples two forms of adaptation by updating an unknown object's dynamics while refining the long-horizon plan.

Overall, \method integrates meta-learning and online dynamics adaptation with adaptive kinodynamic planning within a unified manipulation framework. The meta-learned model provides transferable estimates for previously unseen objects. Meanwhile, the planning tree adapts as the dynamics evolve, retaining previous planning information while staying consistent with the updated model. In the next section, we evaluate the performance of different aspects of the proposed method.

%% file: 5.Experiments.tex
\section{Experiments} 
\label{sec:experiments}
\begin{figure*}[t]
    \centering
    \includegraphics[width=1\textwidth]{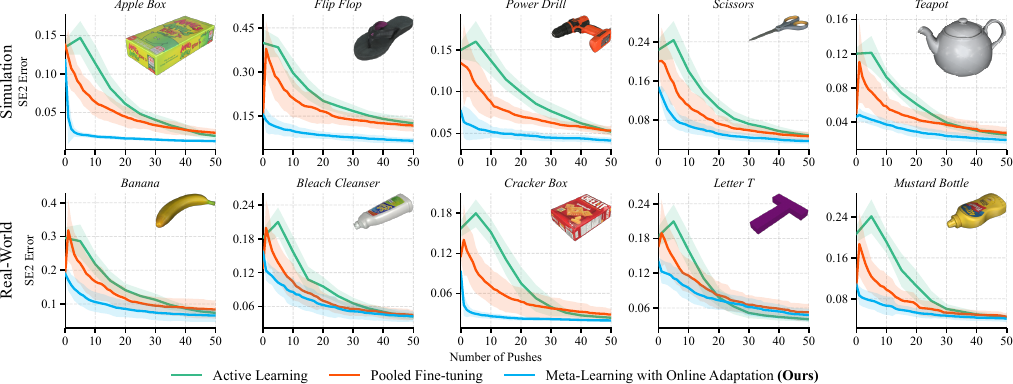}
    \caption{\textbf{Adaptation Performance on Five Simulated and Five Physical Objects.} Prediction error in \SETwo is evaluated on a held-out set, as additional object interactions are used for adaptation (in batches of 10 for \al). For each method, curves are averaged across 30 independent trials, with shaded regions representing one standard deviation.}
    \label{fig:adaptation_comparison}
    \vspace{-1em}
\end{figure*}
We evaluate \method through both dynamics adaptation and manipulation tasks in both simulation and real-world settings. \autoref{subsec:learning} investigates how efficiently the meta-learned dynamics model adapts to previously unseen objects using a few interactions, and \autoref{subsec:manipulation} evaluates how the proposed framework performs in a manipulation task. Moreover, \autoref{subsec:analysis} investigates the effect of changing the training set $\Objs_\train$.

We train the dynamics models in simulation using 238 objects drawn from the YCB Object dataset~\cite{ycb} and Google Scanned Objects~\cite{google}, collecting 2000 simulated pushing transitions for each. For planning, we use \aorrt~\cite{aorrt}
\footnote{The discrete contact point representation in \autoref{sec:methodology} does not satisfy the assumptions required for asymptotic optimality of \aorrt. Although the planner can generate a feasible long-horizon trajectory, in theory, the contact point can be chosen from a continuous boundary of the object.} 
implemented in OMPL~\cite{ompl}. Our setup includes a \mbox{6-Dof UR16e} robot equipped with a rigid cylindrical pushing end effector. For the physical setup, we use the Foundation-Pose model~\cite{foundation-pose} to estimate object pose with an \textsc{I
ntel RealSense D435} depth camera mounted on the end effector.

\begin{figure}[h]
    \centering
    \includegraphics[width=0.85\linewidth]{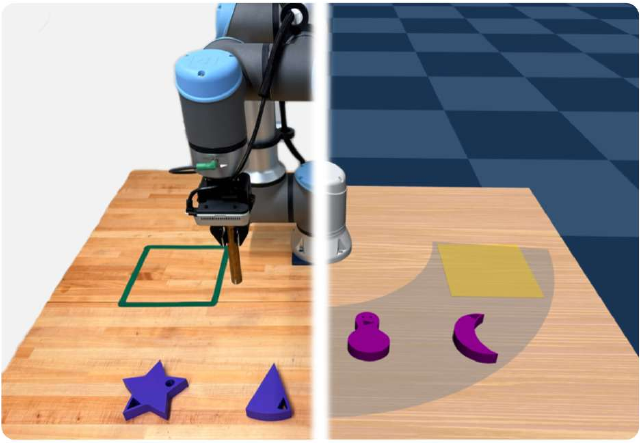}
    \caption{\textbf{Physical and Simulation Setup for Evaluation.} The \textcolor{gray}{gray} annular region shows $\X_\free$ based on the robot's safe workspace, the \textcolor{Dandelion}{yellow region} denotes the start region, and the goal region is on the opposite side in \textcolor{OliveGreen}{green}. Moreover, the planner considers clearance as the safety margin, and MPPI penalizes trajectories that approach the boundaries.}
    \label{fig:env}
    \vspace{-1em}
\end{figure}

Planning is performed in $\X_\free \subseteq \SETwo$, illustrated in \autoref{fig:env} together with the start and goal regions. The object starts at a randomly sampled pose within the start region, and a successful trial is when the object's Oriented Bounding Box (OBB) lies within the goal region. \method is trained according to \autoref{subsec:meta-learning} across the $\Objs_\train$ and is compared against the following baselines:

\begin{itemize}[leftmargin=1.0em]
    \item \textbf{Learning Baselines}
    \begin{itemize}[leftmargin=1.0em]
    \item \textbf{\pft:} Uses the same network architecture, but is trained by pooling all the data from $\Objs_\train$ under a conventional supervised learning objective. This method excludes the meta-learning upside.
    \item \textbf{\al:} Following the active data acquisition strategy of \cite{activepusher}, it actively selects object-specific interactions from the available data pool by acquiring the most informative samples.
    \item \textbf{\hm}~\cite{hacman}: A model-free reinforcement learning method utilizing an actor-critic model, in which the critic learns to pick good contact points, and the actor learns the best action to take at each contact point. Unlike the other baselines, it directly trains a pushing policy instead of learning an explicit pushing model.
    \end{itemize}

    \item \textbf{Planning Baselines}
    \begin{itemize}[leftmargin=1.0em]
        \item \textbf{Learned Model + MPPI~\cite{mppi}:} Serves as a closed-loop controller, which samples candidate action sequences, propagates them through the current learned dynamics model, and evaluates them according to the defined manipulation objective.
    \end{itemize}
\end{itemize}

\subsection{Learning Performance}
\label{subsec:learning}
\begin{figure*}[h]
    \centering
    \includegraphics[width=1.0\linewidth]{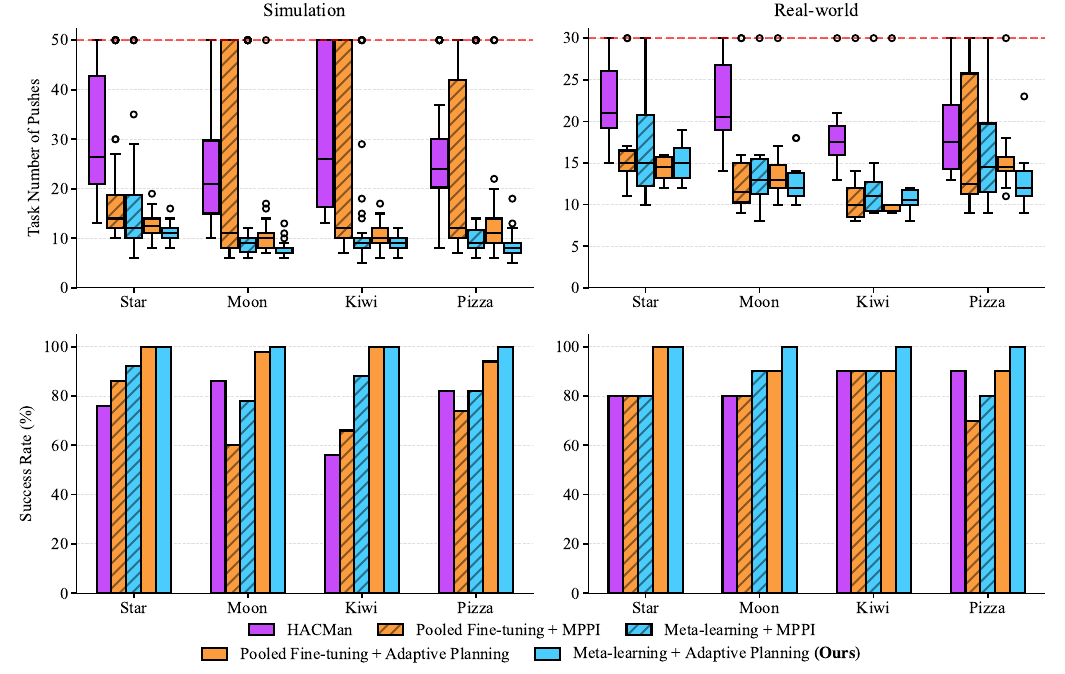}
    \caption{\textbf{Push-to-Goal Manipulation Performance.} Number of pushes required to move the object to the goal across four unseen evaluation objects for 50 trials in simulation and 10 in the real world. Results compare \pft and \ml with MPPI and \ap, together with \hm. A trial fails if the object does not reach the goal within 50 pushes in simulation or 30 pushes in the real-world.}
    \label{fig:task}
    \vspace{-1em}
\end{figure*}
This section evaluates how rapidly each dynamics learning approach improves its predictions when interacting with a new object. In this experiment, we evaluate the dynamics model independently of the downstream manipulation task. We record 50 pushes for 10 different objects in simulation and on physical hardware. In simulation, we collect 2000 independent pushes, reserving 1000 as an evaluation set and using the remaining 1000 as the adaptation pool. For each physical object, the dataset includes 600 pushes, split into 200 for evaluation and 400 for adaptation.

Prediction accuracy is measured on a fixed evaluation set by comparing the predicted and ground-truth state using \autoref{eq:loss-error}. \ml and \pft randomly sample interactions from the adaptation pool set and update the model, whereas \al starts by training on 10 data points and then selects informative interaction batches from the same pool.

As shown in \autoref{fig:adaptation_comparison}, the proposed \ml approach achieves lower prediction error with substantially fewer interactions across both simulation and real-world settings. This method is explicitly trained to provide an initialization that adapts rapidly to unseen objects. In contrast, \pft generally cannot provide reliable predictions during the first few interactions and typically requires more object-specific data to approach comparable prediction accuracy. For some objects, e.g., \textit{Power Drill} and \textit{Banana}, this difference remains visible throughout the evaluated interaction budget.

\al behaves similarly to \pft model. It has a larger prediction error initially, but progressively improves with more interactions. Its accuracy eventually approaches the other two methods for several objects and achieves even lower error for \textit{Letter T}. This demonstrates that \al can produce accurate dynamics with sufficient single-object interactions, however, it trains a separate model from scratch for each new object and is not transferable to different geometries.

Importantly, although the meta-trained dynamics model is trained entirely from simulated push data, it provides an effective initialization for unseen object behavior in the real world. Despite differences in pushing dynamics from the simulated environment to the real world, our learned initialization adapts rapidly with only a limited number of physical interactions, demonstrating the method's sim-to-real transfer capability.

\subsection{Manipulation Performance}
\label{subsec:manipulation}
We next evaluate how online dynamics adaptation and our adaptive planning affect performance on a long-horizon manipulation task. Here, we consider four combinations of a learned model and a controller: \pft with MPPI, \ml with MPPI, \pft with \ap, and \ml with \ap. The final combination corresponds to the complete \method framework.

\autoref{fig:task} compares the number of pushes required to reach the goal and the corresponding task success rate of different methods across the four evaluation objects in simulation and the real world. Overall, the results show that rapid dynamics model adaptation that improves manipulation performance, together with long-horizon planning that can exploit this improved model, yields a more efficient and robust trajectory toward the goal.
The simulation results first demonstrate the effect of dynamics adaptation while keeping the controller fixed. Using MPPI, replacing the \pft model with a \ml model increases the task success rate by \textbf{13.5\%} ($71.5\% \rightarrow 85.0\%$) and decreases the number of pushes required by \textbf{6.38} ($22.52 \rightarrow 16.14$), as faster dynamics adaptation translates into more effective actions. A similar trend holds with \ap, where Meta-learning achieves a success rate of \textbf{100\%} across all trials and reduces the number of pushes by \textbf{2.9} ($11.99 \rightarrow 9.09$). These results are consistent with \autoref{subsec:learning}, i.e., improving the dynamics model at early stages of interaction leads to more reliable predictions for manipulation.

Furthermore, \autoref{fig:task} shows a substantial difference between MPPI and \method's \ap when the dynamics model is held fixed. With \pft, \ap increases task success by \textbf{26.5\%} ($71.5\% \rightarrow 98\%$) compared to MPPI. Using \ml, the proposed planning strategy increases the success rates from \textbf{85\%} for MPPI to \textbf{100\%}. Unlike MPPI, which optimizes actions over a short receding horizon, the planner searches for a dynamically feasible trajectory extending to the goal. Moreover, as described in \autoref{fig:adaptive_planning}, \method adapts the existing planning tree as the learned dynamics evolve, rather than replanning entirely from scratch.

The real-world experiment further evaluates the complete framework under perception noise, execution uncertainty, and sim-to-real dynamics differences. Despite the perception noise, execution uncertainties, and sim-to-real dynamics differences, \method successfully completes the manipulation task in all evaluated trials across all four objects, achieving a \textbf{100\%} success rate.

\hm, as a policy-based baseline, achieves lower success rates and generally requires more pushes than the model-based methods, due to the long-horizon nature of the task and the hard workspace boundaries. \hm's struggles also stem from its inability to adapt to object-specific dynamics models during execution.

\subsection{Training Analysis}
\label{subsec:analysis}
We further evaluate how well \method generalizes to unseen objects by comparing it with models that include evaluation objects during training. This comparison against the \pft baseline here differs from the one in \autoref{subsec:learning} and \autoref{subsec:manipulation}. Here, this model is trained using $\Objs_\train \cap \Objs_\test$, while Object-specific training learns a separate model for each object using interaction data from only that object $o_i \in \Objs_\test$. In contrast, \method is meta-trained exclusively on the $\Objs_\train$ set and has no interaction with $\Objs_\test$ before the task. All methods use \ap, isolating the effect of prior experience on manipulation performance for the evaluation objects.

As shown in \autoref{tab:training_analysis}, \method generalizes well despite not having access to interaction data of any evaluation object. Compared with \pft, which has seen all four objects during training, our method requires fewer pushes per object and achieves consistently lower tracking error. Hence, \method's performance does not rely on including target objects in the training set.

Moreover, the proposed framework achieves performance comparable to Object-specific training, which is trained only on the same object it is evaluated on. Nevertheless, \method achieves lower tracking error for \textit{Mustard~Bottle}~($39.6\%$ improvement), \textit{Letter~T}~($15.8\%$ improvement), and \textit{Banana}~($8.1\%$ improvement), while remaining close on \textit{Cracker~Box}. Both the proposed framework and the \osm succeed on all trials across all four objects, with \method taking fewer pushes for \textit{Mustard~Bottle} and \textit{Banana}.\

Overall, although Object-specific training has direct access to each evaluation object's interaction data, our method can achieve comparable, and in several cases better, performance by adapting the dynamics model online during the task. These results show that the model's meta-learned initialization is not specific to evaluation objects and is transferable to unseen objects.

\begin{table}[t]
\centering
\caption{Task performance across four evaluation objects under different
training-data regimes using \ap.}
\label{tab:training_analysis}

\footnotesize
\setlength{\tabcolsep}{3pt}
\renewcommand{\arraystretch}{1.08}

\begin{tabular*}{\columnwidth}{
@{\extracolsep{\fill}}
l
ccc
@{}
}
\toprule

\textbf{Object, {$\mathbf{i}$}}
& \makecell{\textbf{\textsc{Pooled}\xspace}\\\textbf{\textsc{Fine-tuning}\xspace}}
& \makecell{\textbf{\method}}
& \makecell{\textbf{\textsc{Object-specific}\xspace}\\\textbf{\textsc{Model}\xspace}} \\

\midrule

\textit{Training data}
& \makecell{$\Objs_\train \cap \Objs_\test$}
& \makecell{$\Objs_\train$}
& \makecell{$o_i$} \\

\midrule

\multicolumn{4}{l}{\textit{\textbf{Cracker Box}}} \\

Success Rate
& \textbf{100\%}
& \textbf{100\%}
& \textbf{100\%} \\

Number of Pushes
& 15.46
& 10.88
& \textbf{9.68} \\

Tracking Error
& 0.0731
& 0.0324
& \textbf{0.0317} \\

\addlinespace[3pt]
\multicolumn{4}{l}{\textit{\textbf{Mustard Bottle}}} \\

Success Rate
& 98\%
& \textbf{100\%}
& \textbf{100\%} \\

Number of Pushes
& 14.10
& \textbf{9.52}
& 10.36 \\

Tracking Error
& 0.0934
& \textbf{0.0574}
& 0.0950 \\

\addlinespace[3pt]
\multicolumn{4}{l}{\textit{\textbf{Letter T}}} \\

Success Rate
& \textbf{100\%}
& \textbf{100\%}
& \textbf{100\%} \\

Number of Pushes
& 12.52
& 12.14
& \textbf{10.00} \\

Tracking Error
& 0.0789
& \textbf{0.0545}
& 0.0647 \\

\addlinespace[3pt]
\multicolumn{4}{l}{\textit{\textbf{Banana}}} \\

Success Rate
& 98\%
& \textbf{100\%}
& \textbf{100\%} \\

Number of Pushes
& 14.02
& \textbf{11.56}
& 11.96 \\

Tracking Error
& 0.101
& \textbf{0.068}
& 0.074 \\

\bottomrule
\end{tabular*}
\end{table}

%% file: 6.Conclusion.tex
\section{Conclusion}
\label{sec:conclusion}
%









In this work, we present \method, a framework for nonprehensile manipulation of previously unseen objects in simulation and the real world. \method combines a geometry-based forward dynamics model, meta-training across diverse training objects, online dynamics adaptation from task-generated interactions, and adaptive kinodynamic planning. The meta-learned initialization enables the dynamics model to rapidly adapt to previously unseen objects from limited interaction data, while the planner retains and refines its existing search tree as the dynamics evolve.

\method achieves lower prediction error than \pft and \al by combining \ml with \ap. For a long-horizon manipulation task, it improves task success rate and reduces the number of pushes compared with MPPI-based alternatives. Additionally, \method demonstrates effective sim-to-real transfer by successfully completing all manipulation trials. Furthermore, results show that this method is not object-specific and can transfer to unseen objects while progressively incorporating their latent physical behavior during task execution.

\section{Limitations and Future Work}
\label{sec:limitations}
The current framework has several limitations. The dynamics model relies on accurate contour point cloud observations and state estimation. Moreover, the current formulation considers planar manipulation in $\SETwo$ only by applying normal-direction push actions.

Future work will extend the framework to full $\SEThree$ manipulation and more complex contact interactions, and manipulation with multiple contact points. Incorporating force sensing at the end effector could also provide direct information about contact forces and friction. 